\documentclass[sn-mathphys-num]{sn-jnl}

\usepackage{subfig}
\usepackage{multirow}%
\usepackage{amsmath,amssymb,amsfonts}%
\usepackage{amsthm}%
\usepackage{mathrsfs}%
\usepackage[title]{appendix}%
\usepackage{textcomp}%
\usepackage{manyfoot}%
\usepackage{booktabs}%
\usepackage{algorithm}%
\usepackage{algorithmicx}%
\usepackage{algpseudocode}%
\usepackage{listings}%
\usepackage{booktabs}
\usepackage{siunitx}
\usepackage{graphicx}%
\usepackage[table,xcdraw]{xcolor}

\usepackage{mathtools}
\usepackage{amsmath}
\usepackage{amssymb}

\theoremstyle{thmstyleone}%
\theoremstyle{thmstyletwo}%

\theoremstyle{thmstylethree}%

\begin{document}

\title[Surgical Neural Radiance Fields from One Image]{Surgical Neural Radiance Fields from One Image}


\author*[1,2,4]{\fnm{Alberto} \sur{Neri}}\email{alberto.neri@iit.it}
\author[3,4]{\fnm{Maximilan} \sur{Fehrentz}}
\author[1]{\fnm{Veronica} \sur{Penza}}
\author[1]{\fnm{Leonardo} \sur{S. Mattos}}
\author[4]{\fnm{Nazim} \sur{Haouchine}}

\affil[1]{\orgdiv{Biomedical Robotics Lab, Advanced Robotics}, \orgname{Istituto Italiano di Tecnologia}, \city{Genoa}, \country{Italy}}

\affil[2]{\orgdiv{Department of Computer Science, Bioengineering, Robotics and Systems Engineering (DIBRIS)}, \orgname{University of Genoa}, \city{Genova}, \country{Italy}}

\affil[3]{\orgname{Technical University of Munich}, \city{Munich}, \country{Germany}}

\affil[4]{\orgname{Harvard Medical School, Brigham and Women’s Hospital}, \city{Boston}, \country{USA}}

\abstract{ 
\textbf{Purpose}: Neural Radiance Fields (NeRF) offer exceptional capabilities for 3D reconstruction and view synthesis, yet their reliance on extensive multi-view data limits their application in surgical intraoperative settings where only limited data is available. In particular, collecting such extensive data intraoperatively is impractical due to time constraints. This work addresses this challenge by leveraging a single intraoperative image and preoperative data to train NeRF efficiently for surgical scenarios.

\textbf{Methods}: We leverage preoperative MRI data to define the set of camera viewpoints and images needed for robust and unobstructed training. Intraoperatively, the appearance of the surgical image is transferred to the pre-constructed training set through neural style transfer, specifically combining WTC\textsuperscript{2} and STROTSS to prevent over-stylization. This process enables the creation of a dataset for instant and fast single-image NeRF training. 

\textbf{Results}:  The method is evaluated with four clinical neurosurgical cases. Quantitative comparisons to NeRF models trained on real surgical microscope images demonstrate strong synthesis agreement, with similarity metrics indicating high reconstruction fidelity and stylistic alignment. When compared with ground truth, our method demonstrates high structural similarity, confirming good reconstruction quality and texture preservation. 

\textbf{Conclusion}: Our approach demonstrates the feasibility of single-image NeRF training in surgical settings, overcoming the limitations of traditional multi-view methods. 
    }

\maketitle


\section{Introduction and Related Works}
\label{sec1}
Neural Radiance Fields (NeRF) \cite{mildenhall2021nerf} have emerged as robust techniques for 3D view synthesis, reconstruction and registration, gaining traction in surgical applications \cite{wang2022neural}\cite{fehrentz2024intraoperative}. While advancements like InstantNGP \cite{muller2022instant} enable rapid NeRF generation, compatible with surgical guidance requirements, they rely on acquiring multiple images from diverse viewpoints during surgery—a process prone to user-dependent variability and potential errors. In this paper, we propose to generate a NeRF using a single surgical image.

Single-image NeRF has been addressed with different approaches \cite{pixelNeRF}\cite{jain2021putting}\cite{gu2023nerfdiff}\cite{xu2022sinnerf} outside of surgery. For instance, PixelNeRF \cite{pixelNeRF} employs a ResNet-34 encoder and a feature-conditioned NeRF trained on a multi-view dataset of similar scenes to enable single- or few-shot view synthesis. Since PixelNeRF is trained on ShapeNet \cite{chang2015shapenet}, it excels at synthesizing images of objects within similar categories but struggles to generalize to surgical environments. DietNeRF \cite{jain2021putting} uses prior knowledge from a pre-trained image encoder (CLIP) to guide the NeRF optimization process in the few-shot setting. In their single-view setting, they fine-tune PixelNerf synthesis by augmenting the reconstruction loss with a semantic consistency loss derived from the pre-trained CLIP. This additional supervision enables realistic novel view synthesis and plausible completion of unobserved regions even with few input images.
NerfDiff \cite{gu2023nerfdiff} is also trained on large datasets containing multiple scenes, each with at least two views. During training, it jointly optimizes an image-conditioned NeRF and a 3D-aware conditional diffusion model (CDM) across a collection of scenes. However, at test time, it fine-tunes the NeRF using a single input image by generating and refining virtual views with the CDM. 
Finally, SinNeRF \cite{xu2022sinnerf} is the only model reconstructing scenes from a single RGB and depth input without multi-view pre-training. 
The authors apply geometric and semantic supervision to enable realistic renderings of unseen views. Depth information from the reference view is propagated to other viewpoints through image warping, generating geometry pseudo-labels that enforce multi-view consistency. Additionally, a pre-trained Visual Transformer network ensures appearance consistency across views through semantic supervision.
Overall, these methods suffer from higher computational overhead than InstantNGP. Additionally, apart from SinNeRF, which depends on depth information, the others rely on shape priors from large datasets of common objects, limiting their ability to generalize to surgical scenes. 

\vspace{1em}

\noindent \textbf{Contribution.} We propose a novel approach illustrated in Fig. \ref{fig:overview} that requires only a single intraoperative image to train a NeRF. Our method leverages preoperative data to define the set of camera viewpoints and images needed for robust and unobstructed training. Intraoperatively, the appearance of the surgical image is transferred to the pre-constructed training set, enabling instant and fast NeRF training. We present preliminary results from four neurosurgery cases and demonstrate the effectiveness of the proposed approach in synthesizing novel surgical views.

\section{Methods}
\label{sec2}

\subsubsection*{Problem Formulation}
\begin{figure}[h!]
    \centering
    \includegraphics[clip, trim=0 415 0 30, width=1\linewidth]{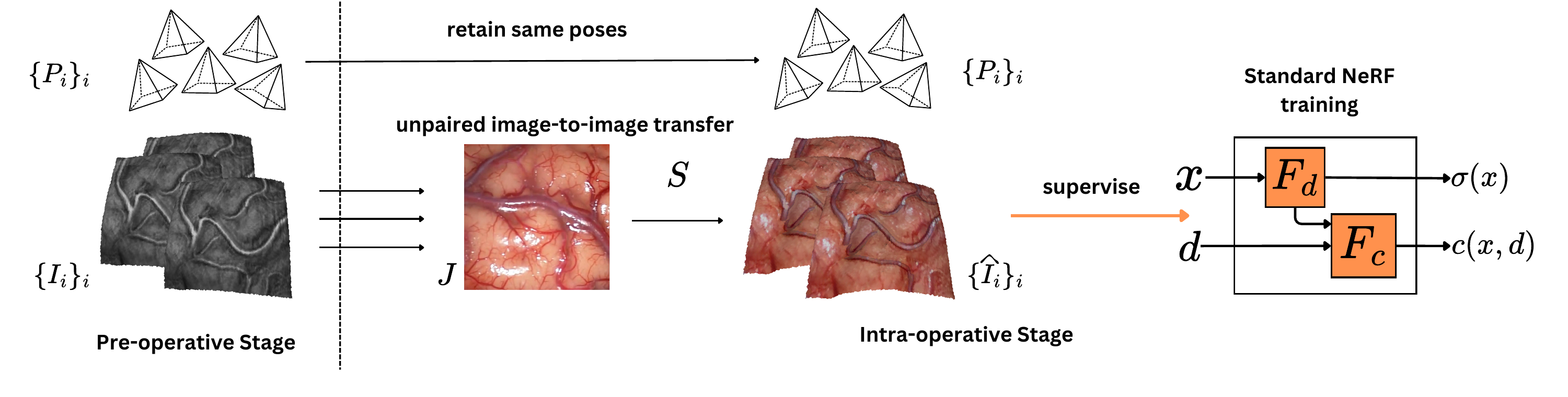}
    \caption{Method overview: the key aspect of our approach is to retain the preoperative poses $\{\mathbf{P}_i\}_i$ and transfer the intraoperative appearance $\mathbf{J}$ to the preoperative images $\{\mathbf{I}_i\}_i$ using $S$, an unpaired image-to-image translation. This process allows for the supervision of the training of a NeRF $F_\theta$ using only one image, circumventing the error-prone multi-view intraoperative acquisition.}
    \label{fig:overview}
\end{figure}

We aim at training a neural radiance field $F_\theta$, where $\theta$ are the learned parameters, that maps spatial positions $\mathbf{x} \in \mathbb{R}^3$ and viewing directions $\mathbf{d} \in \mathbb{S}^2$ to volumetric density $\sigma(\mathbf{x})$ and radiance $\mathbf{c}(\mathbf{x}, \mathbf{d})$ so that  
$F(\mathbf{x}, \mathbf{d}) = \big(\sigma(\mathbf{x}), \mathbf{c}(\mathbf{x}, \mathbf{d})\big)$.
In order to train $F_\theta$, a dataset of intraoperative poses and images $\{(\mathbf{P}_i, \mathbf{J}_i)\}_{i=1}^N$ of $N$ samples is required. 
We want to circumvent this requirement and only use a single intraoperative image $\mathbf{J}$ instead of a multi-view set $\{\mathbf{J}_i)\}_{i=1}^N$, thus lifting the intraoperative burden from surgeon. 
To achieve this we create a training dataset from preoperative scans. 
Using volume rendering to capture brain surface geometry, and by targeting the area where the craniotomy will most likely be placed, we can sample a dataset $\{(\mathbf{P}_i, \mathbf{I}_i)\}_{i=1}^N$ of poses and images.
Because NeRFs are patient-specific, this dataset contains the geometry needed to learn the volumetric density $\sigma(\mathbf{x})$; however, it does not contain the intraoperative appearance to properly train the radiance $\mathbf{c}(\mathbf{x}, \mathbf{d})$.
To this end, we define an unpaired image-to-image transfer function that transfers the intraoperative appearance from $\mathbf{J}$ to the preoperative images dataset $\{\mathbf{I}_i\}_i$.
Our pipeline is illustrated in Fig. \ref{fig:overview}.

\subsubsection*{Intraoperative to Preoperative Appearance Transfer}
Given the intraoperative image $\mathbf{J}$, we want to transfer its appearance to $\{\mathbf{I}_i\}_i$ while retaining the camera poses.
We use Neural Style Transfer (NST) to perform this transformation.
We opted for NST instead of adversarial or variational methods for its deterministic/controlable outputs, fast inference and reduced data requirements. 
Because the preoperative data and intraoperative image are unregistered we need an unpaired image-to-image translation.
In unpaired situations, traditional NST methods do not consistently preserve geometrical consistency and are designed for stylistic rendering at the cost of semantic fidelity. 
We combined WCT\textsuperscript{2}\cite{wct2}, a wavelet-corrected transfer method that preserves the semantic content of images while maintaining photorealism, with STROTSS\cite{strotss}, which utilizes relaxed optimal transport and self-similarity for stylization with controllable guidance. This hybrid approach results in a more balanced style transfer that avoids over-stylization and is performed in two stages. First, we generate an intermediate image, \( \hat{\mathbf{I}}_\text{int} \), using Haar wavelet pooling and whitening/coloring transforms applied in the VGG feature space while minimizing the WCT\textsuperscript{2} loss, \( \mathcal{L}_{\text{WCT\textsuperscript{2}}} \). This stage ensures content preservation and edge detail retention. Then, we apply STROTSS to refine \( \hat{\mathbf{I}}_\text{int} \) for enhanced style fidelity using the loss function \( \mathcal{L}_{\text{STROTSS}} \), which primarily relies on a relaxed Earth Mover’s Distance to match feature statistics and color distributions, leading to a more cohesive and realistic result. The full generation can be formulated as follows:
\begin{equation}
\hat{\mathbf{I}} = \arg\min_{\hat{\mathbf{I}}} \mathcal{L}_{\text{STROTSS}}(\hat{\mathbf{I}}_\text{int}, \mathbf{J}), \quad \text{where} \quad \hat{\mathbf{I}}_\text{int} = \arg\min_{\gamma} \mathcal{L}_{\text{WCT\textsuperscript{2}}}(\mathbf{I}, \mathbf{J}, \gamma)
\end{equation}
This two-stage process is defined by the function $S$ that generates a new image $\hat{\mathbf{I}}$ given the intraoperative image $\mathbf{J}$ and preoperative image $\mathbf{I}$.

\subsubsection*{NeRF Optimization and Training}
In order to train $F_\theta$, we can express the problem so that the rendered pixel intensities approximate the target pixel intensities of the intraoperative image.
This amounts to optimizing the following loss:
\begin{equation}
\mathcal{L}(\theta) = \frac{1}{N} \sum_{i=1}^N \frac{1}{|\Omega|} \sum_{\mathbf{p} \in \Omega} \left\| S(\mathbf{I}_i, \mathbf{J})(\mathbf{p}) - \mathbf{I}_\text{rendered}(\mathbf{p}; \mathbf{P}_i, F_\theta) \right\|^2
\end{equation}
where $\mathbf{p}$ is 2D pixel coordinates in the image $\hat{\mathbf{I}}_i$ and $\mathbf{I}_\text{rendered}(\mathbf{p}; P_i, F_\theta)$ are the color value predicted for pixel $\mathbf{p}$.
We can notice here that we only use a single image $\mathbf{J}$ and that we use the pre-defined set of poses $\{\mathbf{P}_i\}_{i}$, removing the need for multiple intraoperative acquisition. 

To meet the real-time intraoperative requirement, we chose to train NeRF using InstantNGP~\cite{muller2022instant} for computational efficiency. 
It leverages a multi-resolution hash grid encoding to represent the 3D scene efficiently, mapping spatial coordinates and viewing directions to a compact latent space. This latent representation is then passed to a small neural network to predict the color and density values. 
Intraoperatively, the training lasts approximately 150s, while the style transfer takes around 30s. The generation of preoperative poses and images, which occurs in the preoperative phase, takes about 5 minutes. Both style transfer and pose generation can be further optimized.

\section{Experiments and Results}\label{sec3}
\textbf{Dataset and metrics.}
We evaluated our method on 4 clinical neurosurgical cases, each consisting of a preoperative T1 MRI scan and a corresponding surgical microscope image. Each microscopic image was acquired intraoperatively using a standard clinical microscope setup. 
For each case, we generated 100 images, along with their respective camera poses from the preoperative T1 MRI. We use volume rendering to obtain the visualization of the surface of the brain. Then, guided by the surgeon's input on the craniotomy area, we retained only the most relevant portion of the preoperative surface. Thus, we selected camera poses to mimic intraoperative viewpoints realistically.

We used the following performance metrics: the Structural Similarity Index Measure (SSIM) and Peak Signal-to-Noise Ratio (PSNR) for assessing image quality; the Learned Perceptual Image Patch Similarity (LPIPS), commonly used for evaluating synthesis methods; and the Gram Matrix Score (GMS), which measures texture similarity and is widely used in NST approaches. \\

\begin{figure}[h!]
\subfloat{\includegraphics[width=0.16\linewidth]{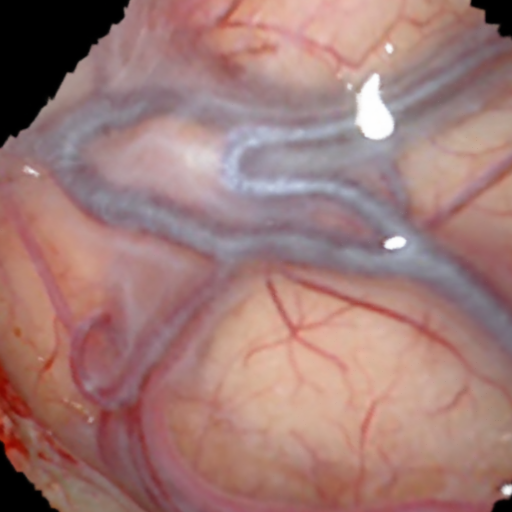}}
\hfill
\subfloat{\includegraphics[width=0.16\linewidth]{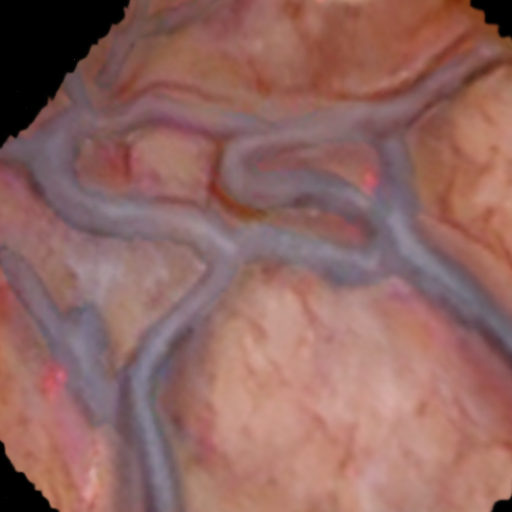}}
\hfill
\subfloat{\includegraphics[width=0.16\linewidth]{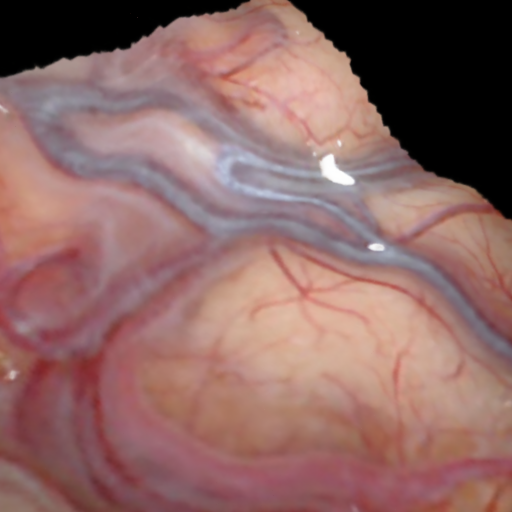}}
\hfill
\subfloat{\includegraphics[width=0.16\linewidth]{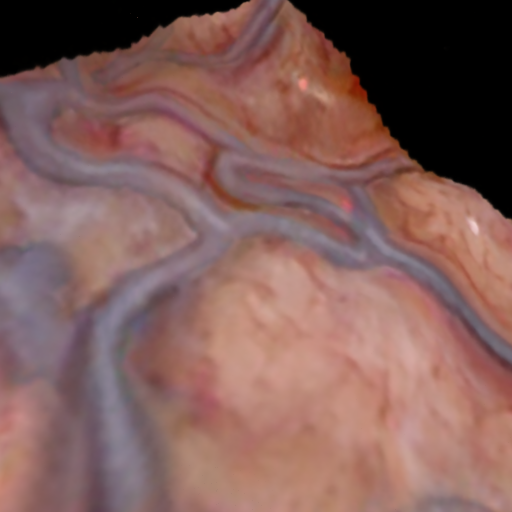}}
\hfill
\subfloat{\includegraphics[width=0.16\linewidth]{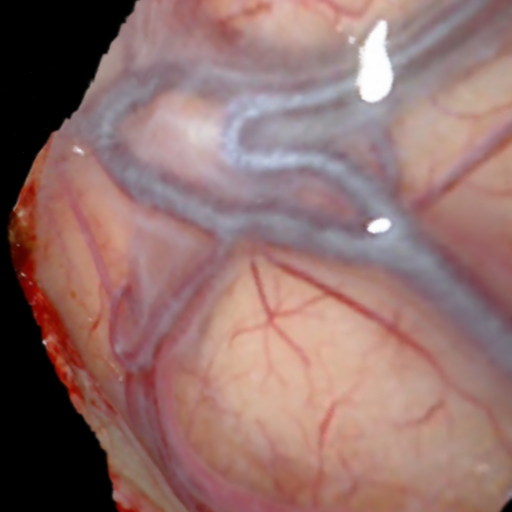}}
\hfill
\subfloat{\includegraphics[width=0.16\linewidth]{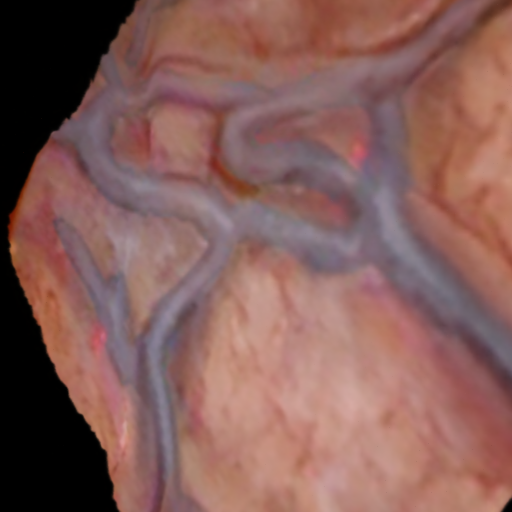}}\\
\hspace{-1.5em}
\subfloat{\includegraphics[clip, trim=100 100 100 100, width=0.16\linewidth]{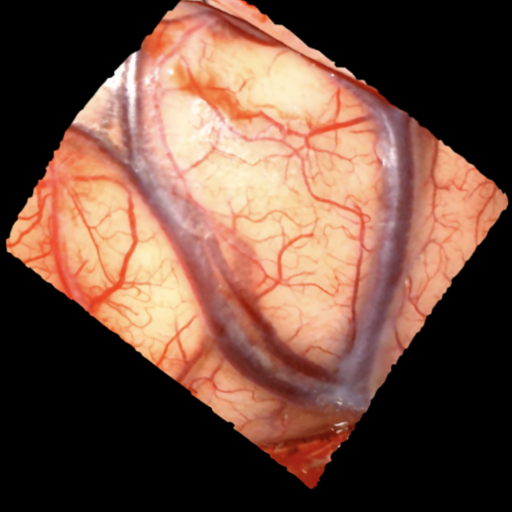}}
\hfill
\subfloat{\includegraphics[clip, trim=100 100 100 100,width=0.16\linewidth]{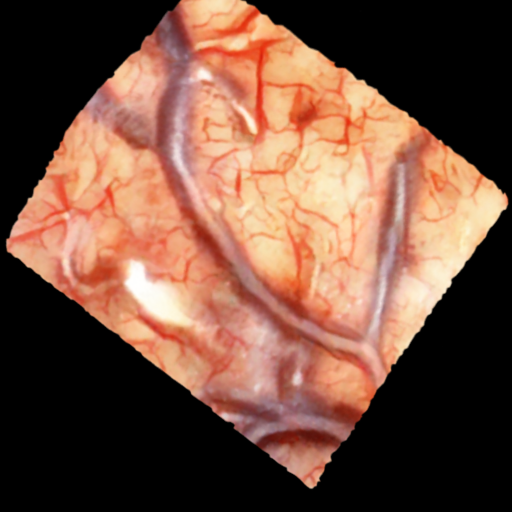}}
\hfill
\subfloat{\includegraphics[clip, trim=150 150 150 150,width=0.16\linewidth]{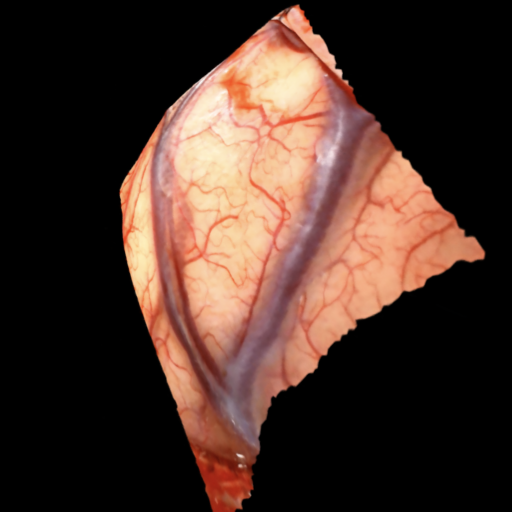}}
\hfill
\subfloat{\includegraphics[clip, trim=150 150 150 150,width=0.16\linewidth]{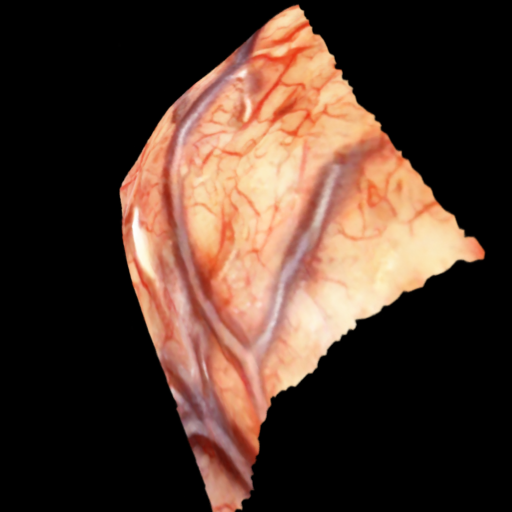}}
\hfill
\subfloat{\includegraphics[clip, trim=100 100 100 100,width=0.16\linewidth]{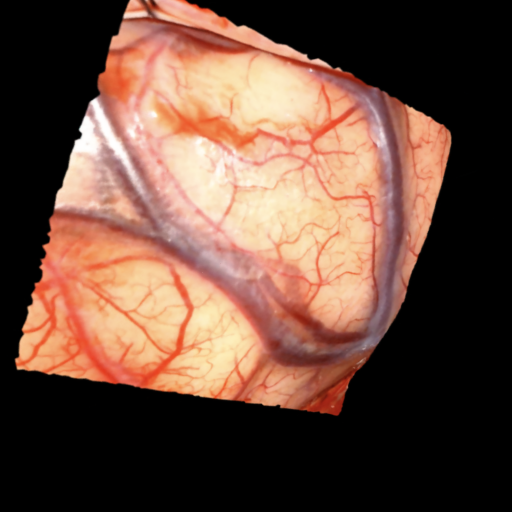}}
\hfill
\subfloat{\includegraphics[clip, trim=100 100 100 100,width=0.16\linewidth]{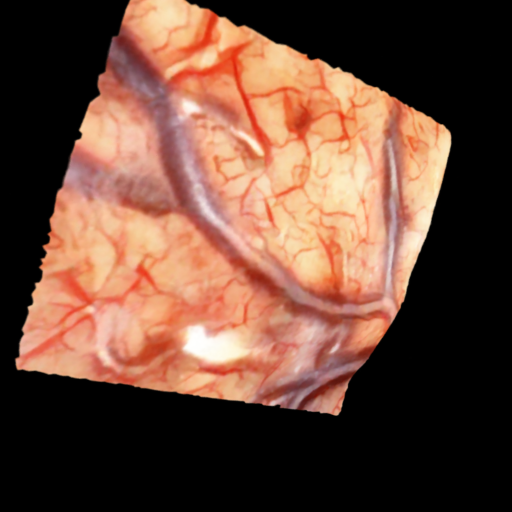}}\\
\hspace{-1.5em}
\subfloat{\includegraphics[clip, trim=100 100 100 100,width=0.16\linewidth]{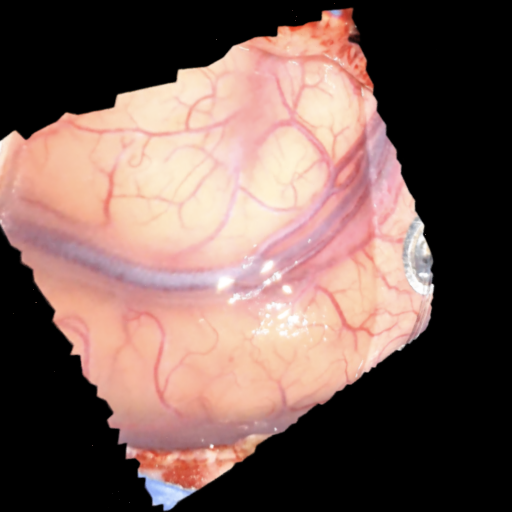}}
\hfill
\subfloat{\includegraphics[clip, trim=100 100 100 100,width=0.16\linewidth]{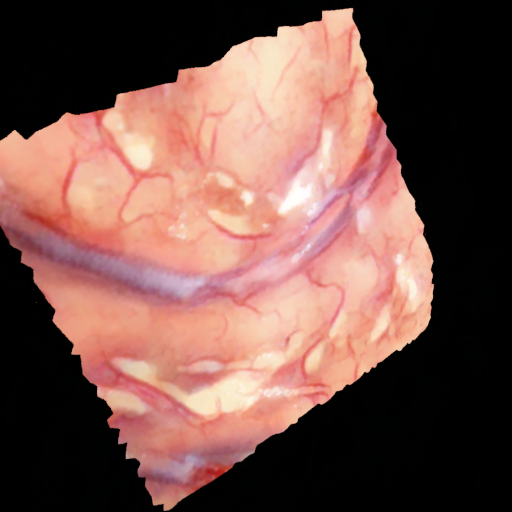}}
\hfill
\subfloat{\includegraphics[clip, trim=100 100 100 100,width=0.16\linewidth]{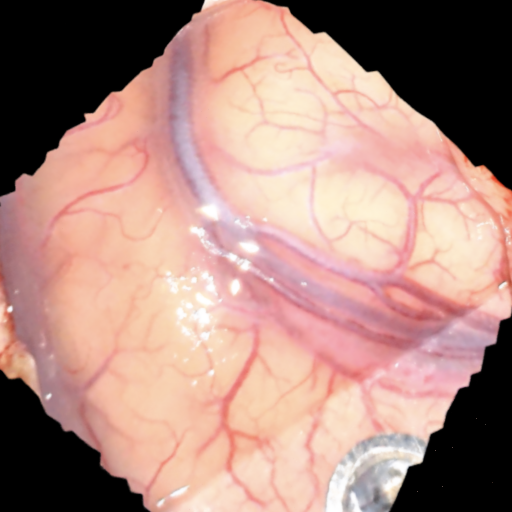}}
\hfill
\subfloat{\includegraphics[clip, trim=100 100 100 100,width=0.16\linewidth]{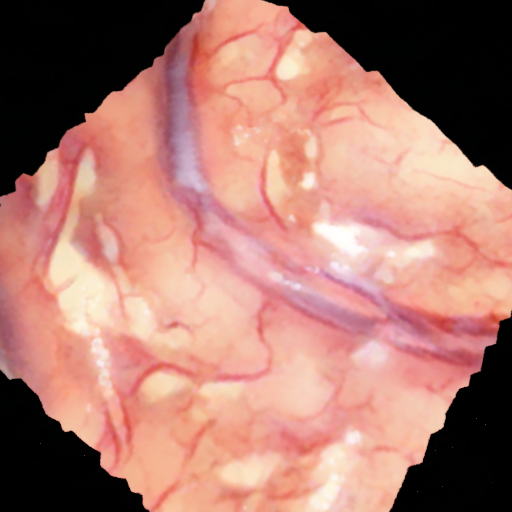}}
\hfill
\subfloat{\includegraphics[clip, trim=100 100 100 100,width=0.16\linewidth]{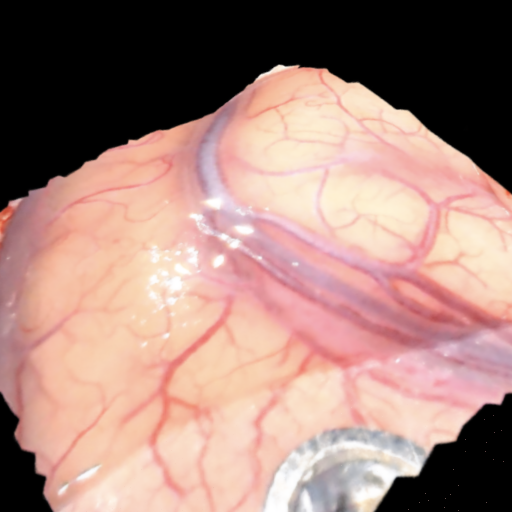}}
\hfill
\subfloat{\includegraphics[clip, trim=100 100 100 100,width=0.16\linewidth]{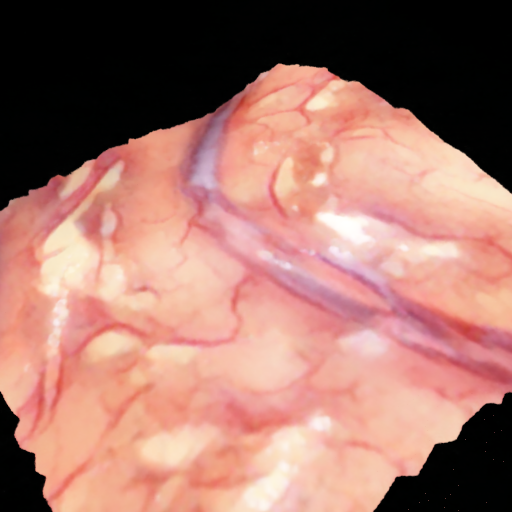}}\\
\hspace{-1.5em}
\subfloat{\includegraphics[clip, trim=100 100 100 100,width=0.16\linewidth]{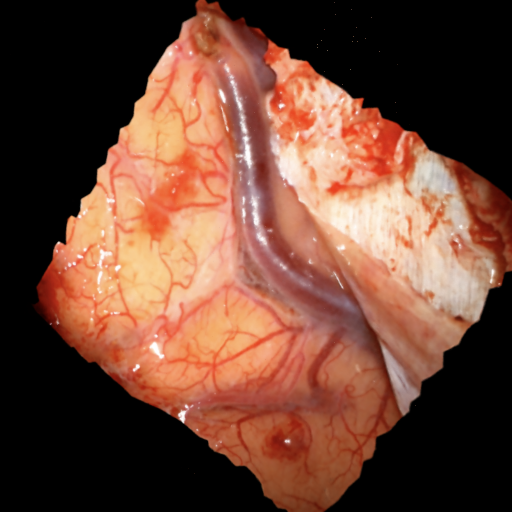}}
\hfill
\subfloat{\includegraphics[clip, trim=100 100 100 100,width=0.16\linewidth]{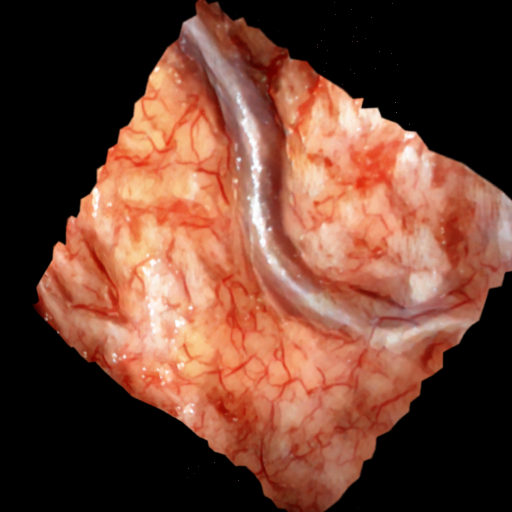}}
\hfill
\subfloat{\includegraphics[clip, trim=100 100 100 100,width=0.16\linewidth]{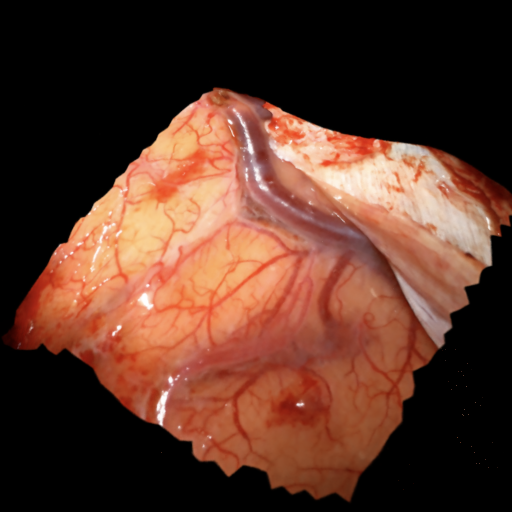}}
\hfill
\subfloat{\includegraphics[clip, trim=100 100 100 100,width=0.16\linewidth]{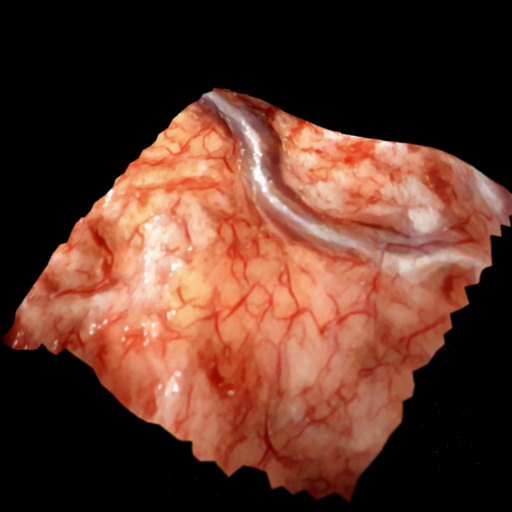}}
\hfill
\subfloat{\includegraphics[clip, trim=100 100 100 100,width=0.16\linewidth]{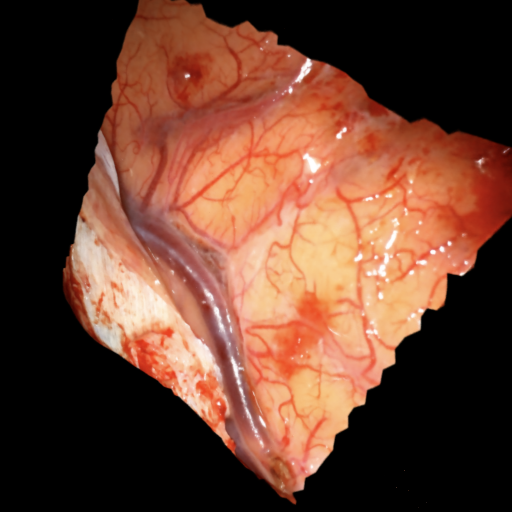}}
\hfill
\subfloat{\includegraphics[clip, trim=100 100 100 100,width=0.16\linewidth]{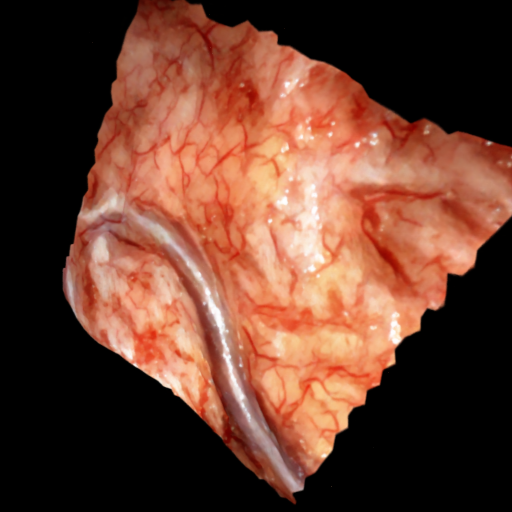}}

\caption{Tests on four real cases, one per row with pairs of synthesized images from three random viewpoints using MV-NeRF and our method, respectively.}
\label{fig:clinical}
\end{figure}

\noindent \textbf{Agreement with multi-view NeRF.}
We trained a NeRF using our method and measured its synthesis agreement with a Multiview NeRF (MV-NeRF) trained on real surgical microscope images.
For evaluation, we synthesized 9 images from random poses to account for the small observed brain region (50 mm diameter). This approach simulates intraoperative viewpoint variability and ensures out-of-distribution testing by avoiding overlap with training poses.
Table \ref{tab:results} shows the average results across all poses for each case.
All experiments achieved SSIM scores above $0.74$ and PSNR values greater than 30dB, indicating that our method consistently maintains the structural integrity of the images. The LPIPS values varied between $0.20$ and $0.34$, suggesting that our method achieves a good balance between perceptual similarity and image quality. 
Lastly, except for \textit{Case 3}, the GMS values remained below $0.15$, indicating that the generated images closely match the MV-NeRF regarding textural representation.
We provide qualitative results with a visual assessment of each case in Fig. \ref{fig:clinical}.

\begin{table*}[h!]
\centering
\caption{Average synthesize agreement with MV-NeRF across all poses.}
\label{tab:results}
\resizebox{0.90\textwidth}{!}
{
\begin{tabular}{lcccc}
\toprule
       & SSIM & PSNR & LPIPS &  GMS \\
\midrule
\rowcolor{gray!10}
\textit{Case 1} &   0.78 $\pm$ 0.02	& 30.26 $\pm$  0.88	&	0.34 $\pm$ 0.06 & 0.10 $\pm$ 0.04  \\
\textit{Case 2} &   0.79 $\pm$ 0.03 & 31.71 $\pm$ 0.89 & 0.30 $\pm$ 0.06 & 0.15 $\pm$ 0.04  \\
\rowcolor{gray!10}
\textit{Case 3} &   0.74 $\pm$ 0.03 & 32.51 $\pm$ 0.70 & 0.23 $\pm$ 0.04 & 0.28  $\pm$ 0.07 \\
\textit{Case 4} &   0.79 $\pm$ 0.04 & 33.39 $\pm$ 0.85	& 0.20 $\pm$ 0.04 &	0.13 $\pm$  0.04  \\  
\bottomrule
\end{tabular}
}
\end{table*}

\begin{figure}[h!]
\subfloat{\includegraphics[clip, trim=10 10 10 40, width=0.249\linewidth]{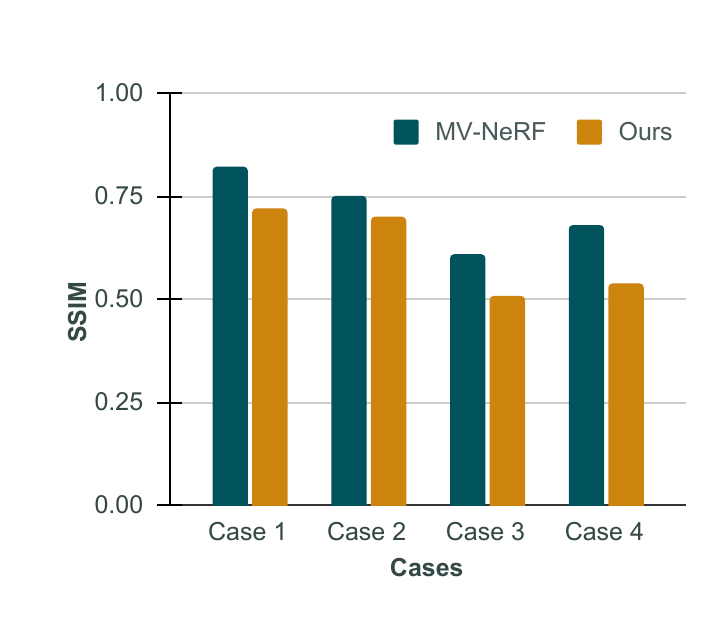}}
\hfill
\subfloat{\includegraphics[clip, trim=10 10 10 40, width=0.249\linewidth]{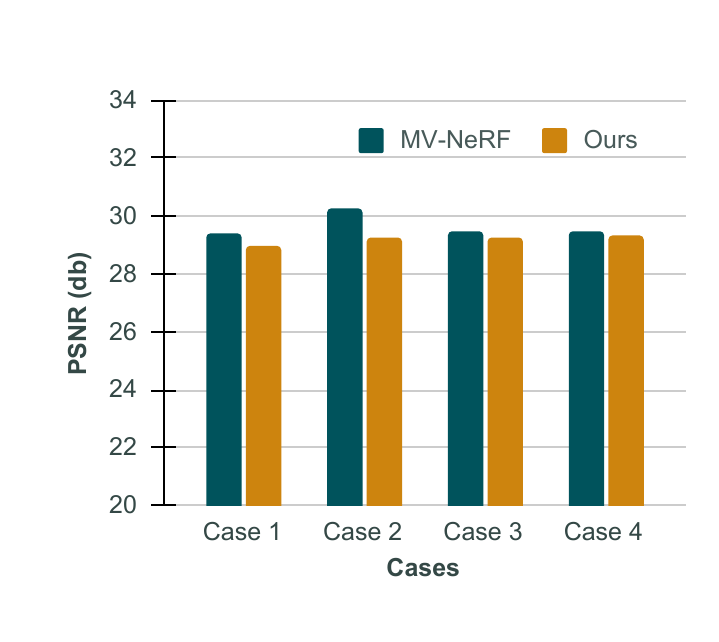}}
\hfill
\subfloat{\includegraphics[clip, trim=10 10 10 40, width=0.249\linewidth]{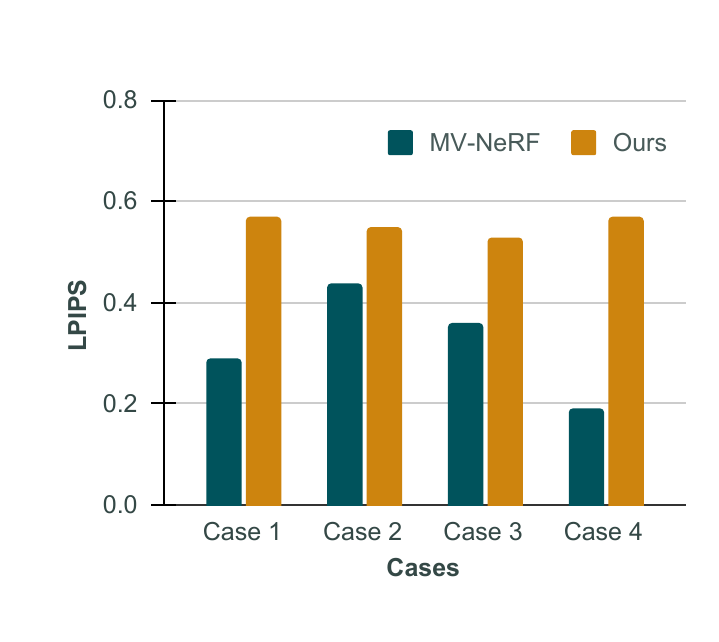}}
\hfill
\subfloat{\includegraphics[clip, trim=10 10 10 40, width=0.249\linewidth]{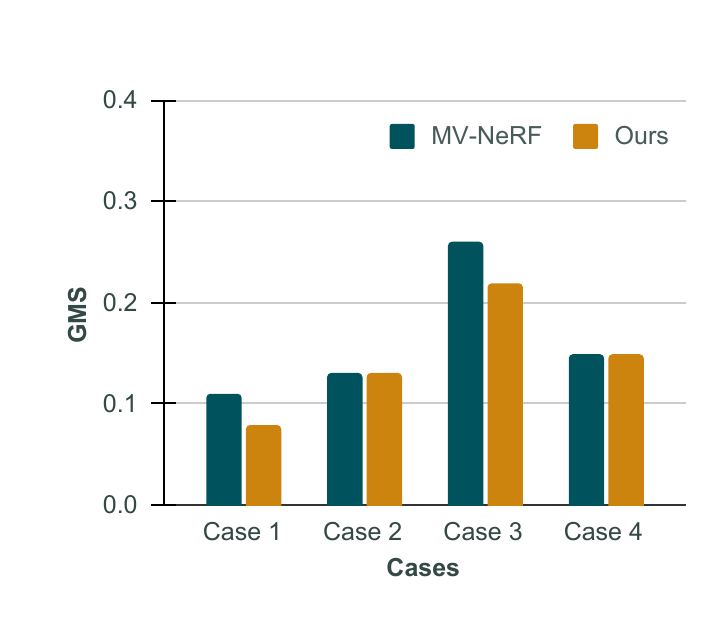}}
\caption{Comparison against ground-truth image after registration.}
\label{fig:results-gt}
\end{figure}

\noindent \textbf{Comparison against ground-truth.}
To compare with the ground-truth surgical image, we manually registered the preoperative MRI's brain surface with the real surgical image to obtain a ground-truth camera position and orientation. 
We used this information to synthesize one image per case using our methods and MV-NeRF and compare them with the real surgical image. 
The plots in Fig. \ref{fig:results-gt} show the results for each case.
In Cases 1 and 2, SSIM exceeds 0.70, and in Cases 3 and 4 it remains above 0.50, likely due to mis-registration. Overall, the SSIM difference between MV-NeRF and our method is less than 10\%. PSNR is consistently around 30dB in all cases, suggesting good reconstruction quality. Although LPIPS indicates some deviations due to the unpaired nature of the NST, GMS confirms that our synthesized images faithfully preserve style and texture—with scores remaining below 0.15, except for Case 3, consistent with the results shown in Table \ref{tab:results}.

\captionsetup[subfigure]{labelformat=empty}

\captionsetup[subfigure]{labelformat=empty}

\begin{figure}[h!]
\centering
\includegraphics[clip, trim=0 0 0 20, width=0.75\linewidth]{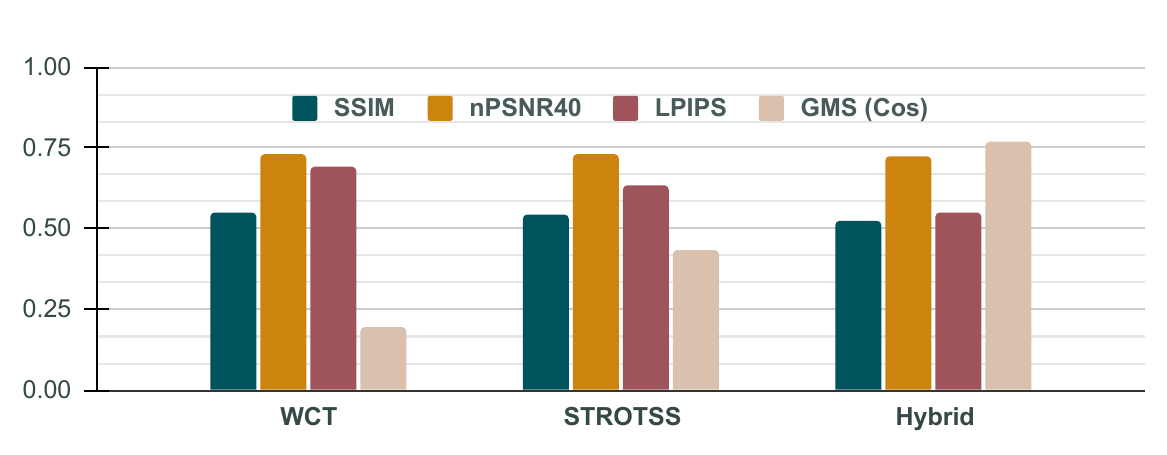}
\caption{Comparison of various NST methods. Our hybrid approach strikes an optimal balance between style and structure. For readability, PSNR and GMS are scaled to the range [0,1]: PSNR is normalized with a maximum value of 40dB, and GMS is computed using cosine similarity.}
\label{fig:results-nst}
\end{figure}

\noindent \textbf{Impact of neural style transfer methods.} 
We compared the impact of the NST methods on image synthesis by measuring the added value of each method when compared with the real image. 
We compared the synthesis using WCT\textsuperscript{2}, STROTSS and the combination of both.
The results reported in Fig. \ref{fig:results-nst} show that our hybrid approach maintains the best balance between style and structure with results comparable to WCT and STROTSS, while outperforming in LPIPS and GMS metrics.

\begin{figure}[h!]
\subfloat{\includegraphics[width=0.24\linewidth]{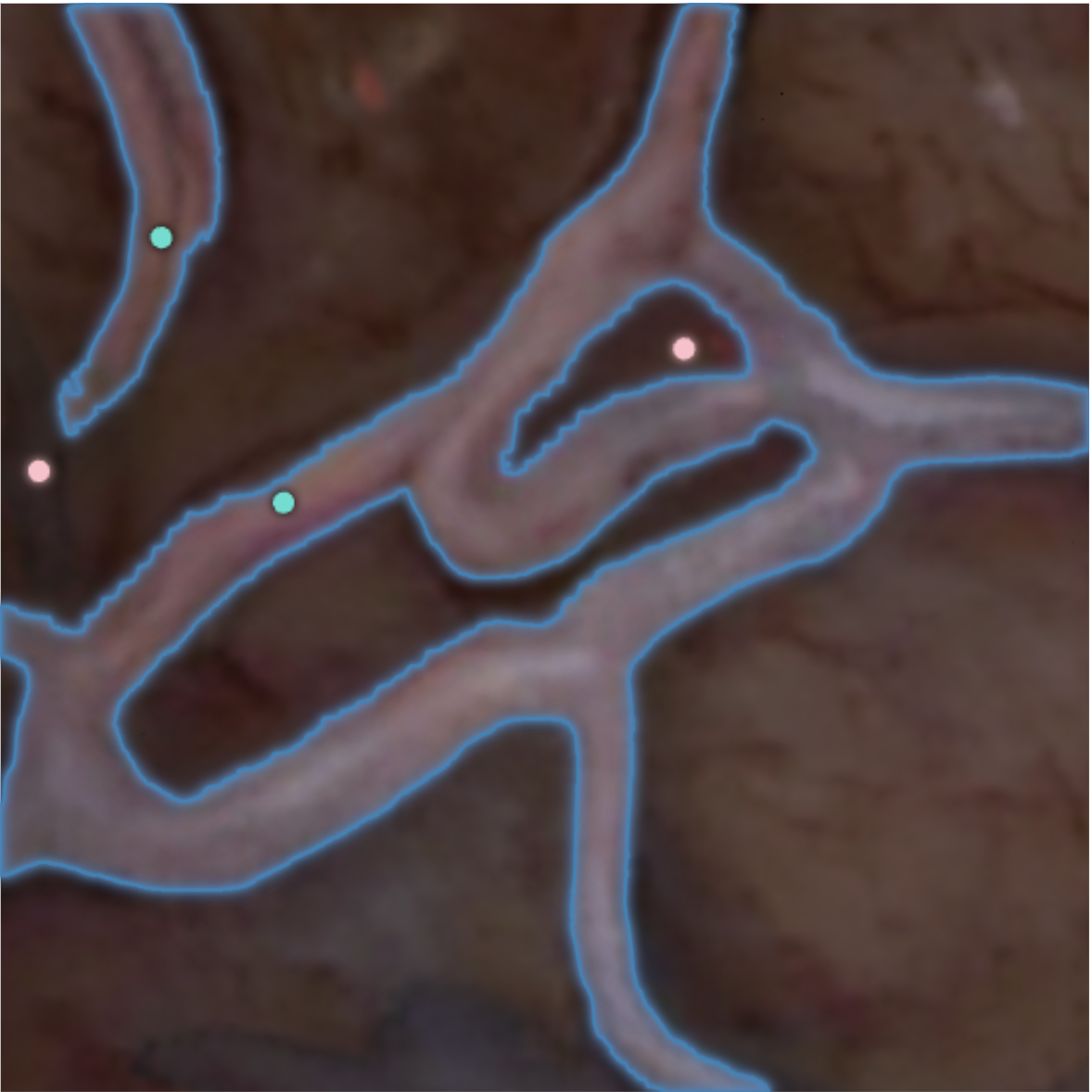}}
\hfill
\subfloat{\includegraphics[width=0.24\linewidth]{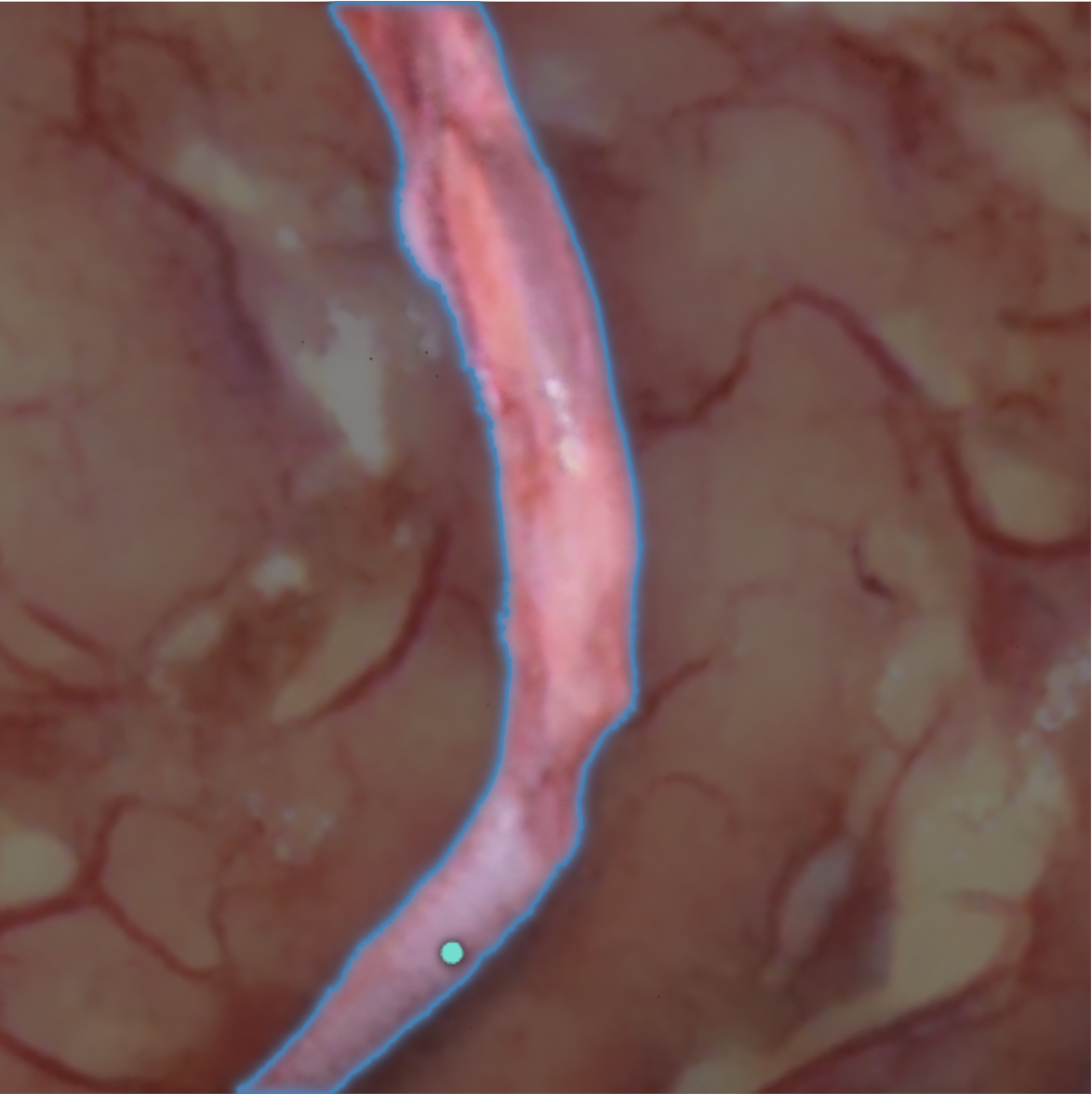}}
\hfill
\subfloat{\includegraphics[width=0.24\linewidth]{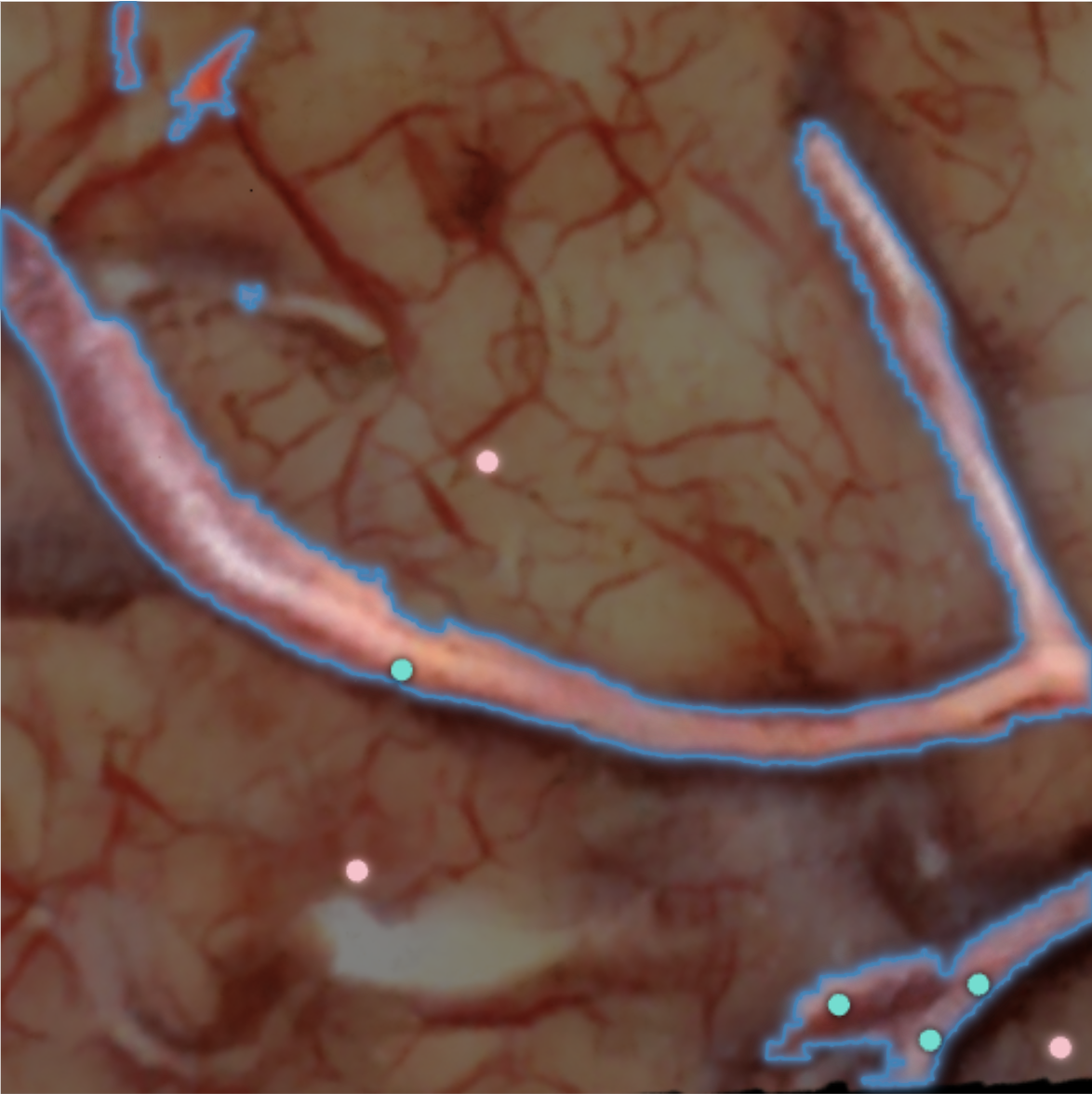}}
\hfill
\subfloat{\includegraphics[width=0.24\linewidth]{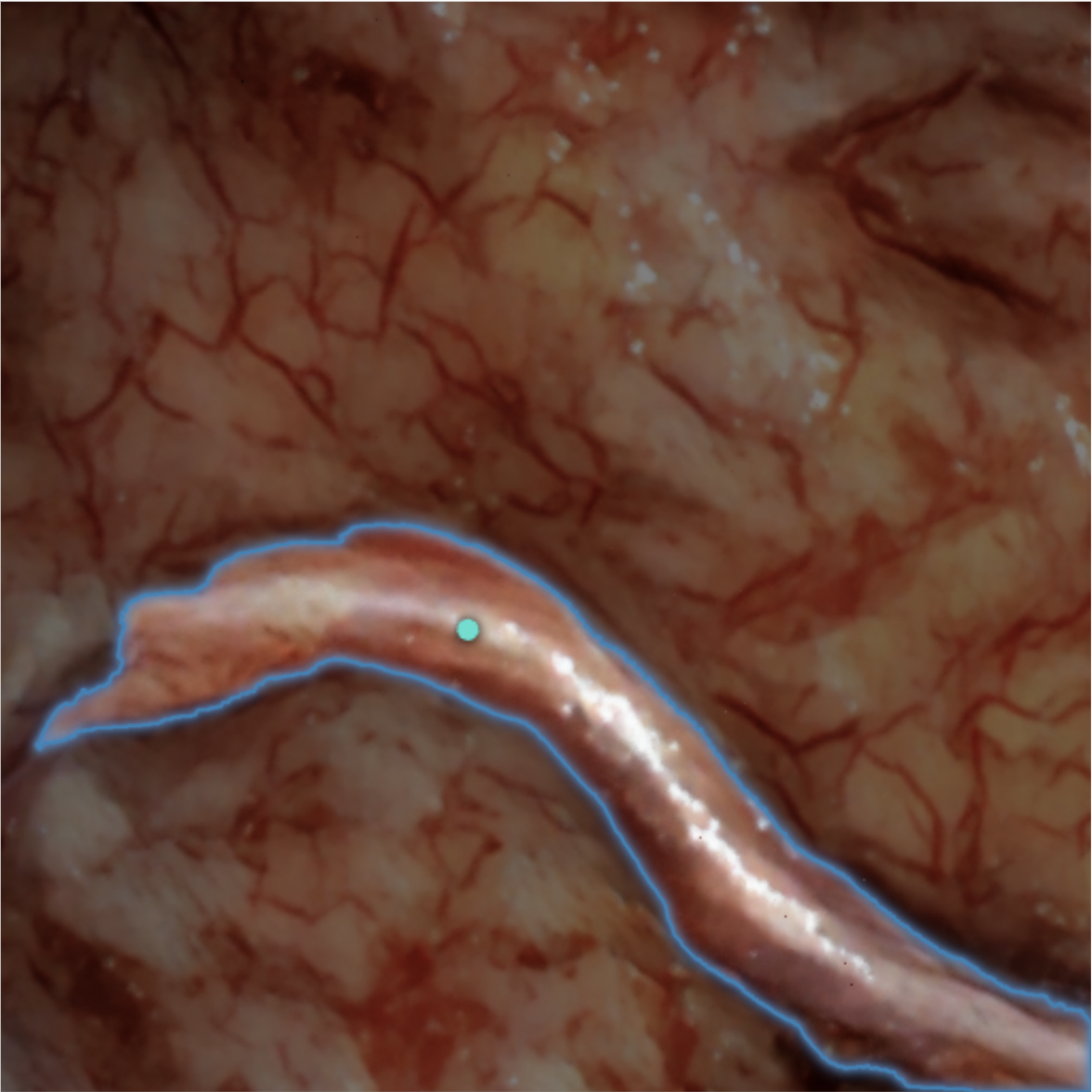}}
\caption{Interactive, semi-automated segmentation on synthesized images.}
\label{fig:seg}
\end{figure}

\noindent \textbf{Interactive segmentation on synthesized images.}
Additionally, we show that our synthesis method can be applied to downstream tasks such as vessel segmentation. 
Using an interactive semi-automated foundation model \cite{SAM}, not fine-tuned on surgical brain dataset, we obtain good segmentation results as illustrated in Fig. \ref{fig:seg}.

\captionsetup[subfigure]{labelformat=empty}

\section{Conclusions}
We introduced a novel approach for intraoperative surgical NeRF training using only a single image. Our preliminary results demonstrate that image synthesis from our single-image NeRF leads to good agreement with a multi-view strategy, and achieves good reconstruction and texture preservation when compared with ground-truth image. 
A limitation of our method is that the intraoperative style is static and does not account for variations in lighting conditions. Future work will focus on validating the method in a downstream registration task where differentiable representations can be used for 3D/2D registration and camera pose estimation.

\section{Declarations}
\textbf{Conflict of interest} This work has no competing interests to declare. 

\bibliography{main}

\end{document}